\documentclass{article}

\usepackage{PRIMEarxiv}

\usepackage[utf8]{inputenc} 
\usepackage[T1]{fontenc}    
\usepackage{hyperref}       
\usepackage{url}            
\usepackage{booktabs}       
\usepackage{amsfonts}       
\usepackage{nicefrac}       
\usepackage{microtype}      
\usepackage{lipsum}
\usepackage{fancyhdr}       
\usepackage{graphicx}       
\graphicspath{{media/}}     
\usepackage{float}
\usepackage{caption}
\usepackage[numbers, sort&compress]{natbib} 

\title{A Case for Leveraging Generative AI to Expand and Enhance Training in the Provision of Mental Health Services

}

\author{
  Hannah R Lawrence \\
  Google \\
  Mountain View, CA, United States\\
  \texttt{hannahlawrence@google.com} \\
   \And
  Shannon Wiltsey Stirman \\
  National Center for PTSD \\
  Stanford University \\
  Stanford, CA, United States\\
   \And
  Samuel Dorison \\
  ReflexAI \\
  New York, NY, United States\\
   \And
  Taedong Yun \\
  Google \\
  Mountain View, CA, United States\\
   \And
  Megan Jones Bell \\
  Google \\
  Mountain View, CA, United States\\
  \texttt{meganjonesbell@google.com} \\
}

\begin{document}
\maketitle

\begin{abstract}
Generative artificial intelligence (Generative AI) is transforming healthcare. With this evolution comes optimism regarding the impact it will have on mental health, as well as concern regarding the risks that come with generative AI operating in the mental health domain. Much of the investment in, and academic and public discourse about, AI-powered solutions for mental health has focused on therapist chatbots. Despite the common assumption that chatbots will be the most impactful application of GenAI to mental health, we make the case here for a lower-risk, high impact use case: leveraging generative AI to enhance and scale training in mental health service provision. We highlight key benefits of using generative AI to help train people to provide mental health services and present a real-world case study in which generative AI improved the training of veterans to support one another’s mental health. With numerous potential applications of generative AI in mental health, we illustrate why we should invest in using generative AI to support training people in mental health service provision.
\end{abstract}


\section{Introduction}
There is a dramatic shortage of mental health providers globally. There are only 13 mental health providers for every 100,000 people worldwide \cite{WorldHealth2020}, resulting in a global treatment gap of over 50

Here, we make a case for instead leveraging generative AI to support training people in high quality mental health service provision. These types of use cases are possible because of recent technical advances in generative AI that have improved performance in similar applications \cite{Steenstra2025}. Often, a Retrieval-Augmented Generation (RAG) architecture is used to provide the AI with a training playbook or handbook, ethical guidelines, and best practice examples. These applications should also undergo testing by domain experts using evaluation rubrics tailored to the use case before deployment. As such, it is not simply “out of the box” generative models with a simple prompt, but rather thoughtfully designed and tested AI models that could make a true difference in expanding and enhancing training in mental health service provision.

Below, we detail specific ways generative AI can be used to improve mental health training. There are opportunities to scale training to substantially increase the mental health workforce. The quality of mental health services can be improved through interactive practice with generative agents acting as clients with a wide range of identities, backgrounds, and presenting concerns. Supervision can be enhanced by providing trainees with in-the-moment feedback and information on strengths, areas for growth, and progress over time. This model of using AI as a learning and development tool but maintaining human-delivered support also has secondary benefits of creating potential employment opportunities as in the case of peer support specialists and aligns with global calls to promote social connection. 

\begin{table}[htbp]
\centering
\caption{Definitions of terms}
\label{my_table}
\begin{tabular}{l|p{4cm}p{6cm}}
\toprule
Term used in this paper & Description & Non-exhaustive examples \\
\midrule
Trainee & Any person engaged in learning skills related to mental health service provision & A clinical psychology doctoral student, a social worker aiming to increase their competence in working with a new population, or a community member learning skills to deliver intervention skills as part of task sharing \\
Clinician & Any person who is engaged in mental health service provision & A clinical psychologist, social worker, psychiatrist, school counselor, primary care physician \\
Client & Anyone that the trainee or clinician is working with clinically & An individual therapy client, a patient presenting to a primary care office with behavioral health concerns, or a person in the community who is seeking mental health support from another trained community member \\
\bottomrule
\end{tabular}
\end{table}

\section{Scalability}
\label{sec:headings}

There is enormous demand for high quality mental health services \cite{Wainberg2017} and increasing clarity that current systems of training and deploying mental health professionals are unlikely to sufficiently scale to meet the need. Despite enormous interest and investment in expanding the mental health workforce globally, insufficient progress has been made in closing this gap between mental healthcare need and mental healthcare access \cite{WorldHealth2020}. This is at least in part because training humans to provide high quality mental health care is costly, time intensive, resource intensive, and limited in moments of increased need (e.g., during times of conflict). 

Generative AI provides one solution with potential to contribute to closing the treatment gap by scaling high quality training in mental health service provision. Generative AI can be used to train new clinicians, to train established clinicians in new or the most up-to-date treatments, to re-train clinicians who are struggling with a specific skill or who would like to expand their practice to a new population, and to train people without formal education in mental healthcare to deliver empirically-supported interventions in their communities (i.e., as part of task shifting and sharing). Example uses of generative AI to scale training in mental health service provision include powering interactive, personalized didactic training, helping to identify specific training needs for individuals or regions, developing simulated practice clients, and offering performance feedback \cite{TherapyTrainer2025}. 

\begin{figure}[H]
  \centering
  \includegraphics[width=0.8\textwidth]{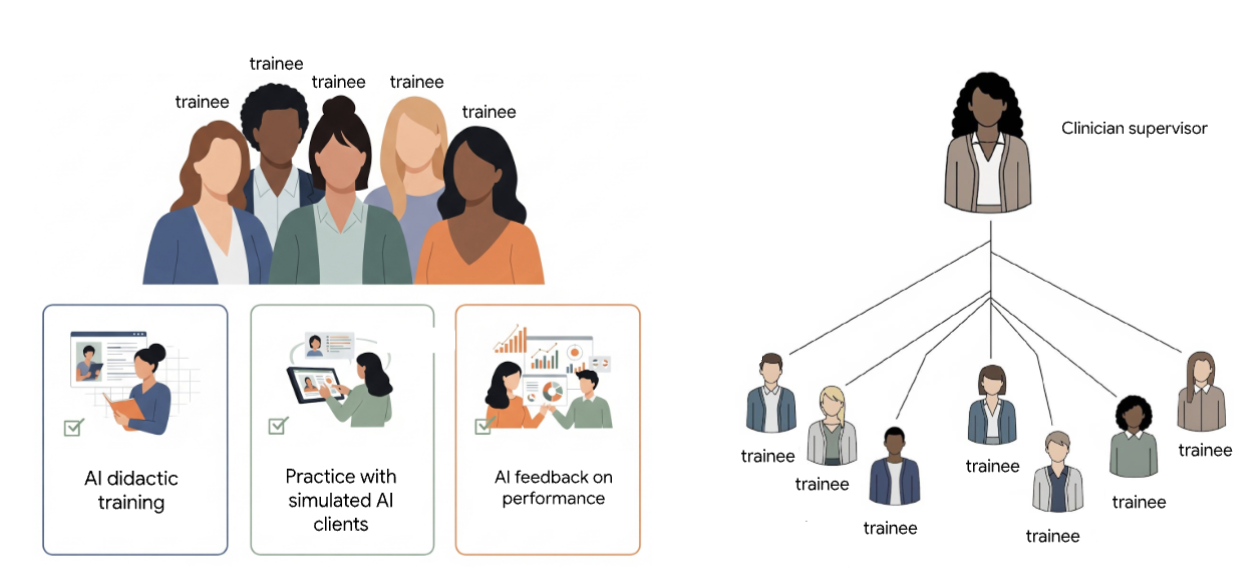}
  \caption{By using AI to deliver personalized didactic training, to generate simulated clients for practice, and to provide feedback on learning and performance during training, there is potential for a single clinician supervisor to focus their full effort on supervision and extend their reach to a greater number of trainees. This figure was made using generative AI. }
  \label{fig:myimage1}
\end{figure}

\section{Opportunities for interactive practice }
There is evidence that people gain more skill in providing mental health services during training when they engage in active practice and role playing \cite{Cross2011, Anderson2020}. The transfer of this learning to work with real people experiencing mental health concerns is limited, however. This is in part due to the fact that role plays rarely resemble the population one will work with in practice. For example, it is common practice for graduate students to participate in role play interactions with other graduate students who pretend to be therapy clients, often before ever seeing clients themselves. Actors who are trained to simulate patient interactions with trainees are often perceived as less realistic, and human simulation of this nature is difficult to coordinate and not feasible for simulation of full courses of mental health treatment \cite{Kuhne2018}. Flexibly practicing therapy skills with simulated clients closely resembling the real clients the trainee will work with can have large positive impacts on trainees' competence.

Generative AI can make learning more active through practice that mimics real life interactions. As one example, generative AI can be used to develop client simulations, with initial research finding that they can be tailored to resemble the real population of people who seek mental health support in a given setting \cite{Chen2023, Louie2024, Demasi2020}. Trainees can practice with these patient simulations, receiving (human or AI-generated) feedback and supervision until they are competent to work with real people. This approach does not put real people in mental health distress at risk by having them work with new trainees, can be scaled and tailored to specific contexts, populations, and clinical presentations, and also provides the opportunity to assess skill acquisition over time. In light of research findings that the first few clients who are treated by clinicians who were recently trained in a new treatment do not do as well as their subsequent clients \cite{Swanson2021, Johnson2019}, simulations of this nature could improve the quality of care and support better outcomes when trainees begin treating clients. Once benchmarks for competency are established, they could also serve a gatekeeping function to ensure adequate therapeutic skills needed prior to interacting with real people.

\begin{figure}[H]
  \centering
  \includegraphics[width=0.8\textwidth]{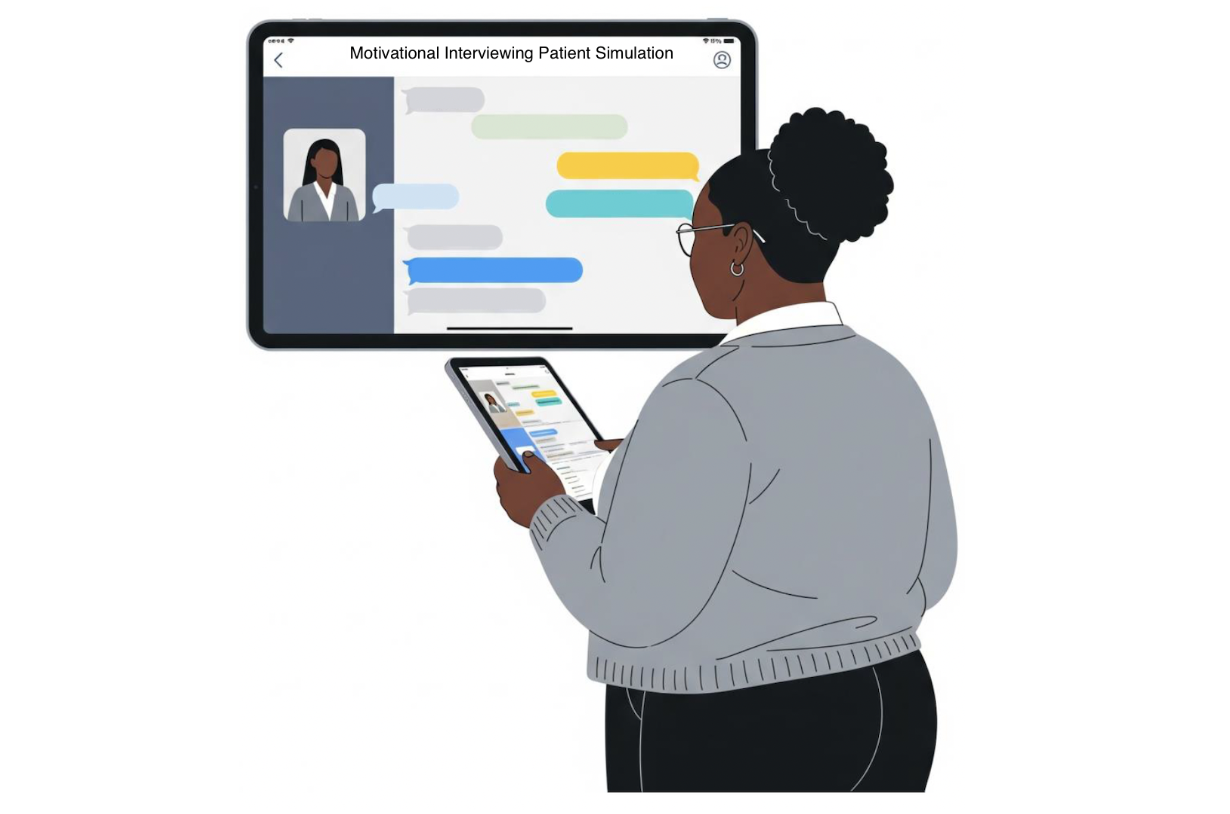}
  \caption{Specific client simulations can be developed to support specific types of skill acquisition or competency development. As one example, a trainee hoping to improve their skills in motivational interviewing could practice with a simulated client experiencing ambivalence about a behavior change, while receiving feedback on their motivational interviewing skills and strengthening competence over time. This figure was made using generative AI. }
  \label{fig:myimage2}
\end{figure}

\section{Opportunities to practice with clients with diverse identities, backgrounds, and presenting concerns}

The clients that trainees typically gain experience working with depends wholly on the client populations of the setting in which they are training. For many, this means exclusively clients in either urban or rural environments (depending on where the training program is located) and clients of racial and ethnic identities that reflect the local population. Additionally, many training environments may not offer clinical services for people presenting with the entire range of mental health concerns or diagnoses. Some clinics specifically serve a certain client population (e.g., an anxiety clinic), others may refer out people with certain concerns (e.g., substance use), and most offer one specific level of care whether it be preventative care, outpatient, partial hospitalization, or inpatient hospitalization (among others).

Generative AI provides the chance to engage with simulated clients that trainees might not otherwise have the opportunity to work with while in training. These clients may present with different cultural backgrounds and beliefs, speak different languages, and present with different or more complex clinical concerns. Simulated client interactions have also been put forward as one way for trainees to practice responding to higher risk or less common clinical situations (e.g., responding to suicide risk) \cite{Sharpless2009}. This would allow trainees to gain experience not only to improve their skill in responding to mental health risk, but also to habituate and regulate one's own emotional reaction to higher risk situations. Opportunities to practice with simulated clients of different backgrounds and with low base rate or more severe mental health concerns while still in training offers trainees the opportunity to receive feedback and gain competence without the risk of harm to real clients, particularly those from marginalized backgrounds or who present with more severe mental health concerns. Importantly, for this to be useful, simulated clients must be carefully developed. Ideally, this would occur with input from actual clients with the backgrounds or presenting concerns that aim to be represented, and from clinicians who regularly work with these populations.

\begin{figure}[H]
  \centering
  \includegraphics[width=0.8\textwidth]{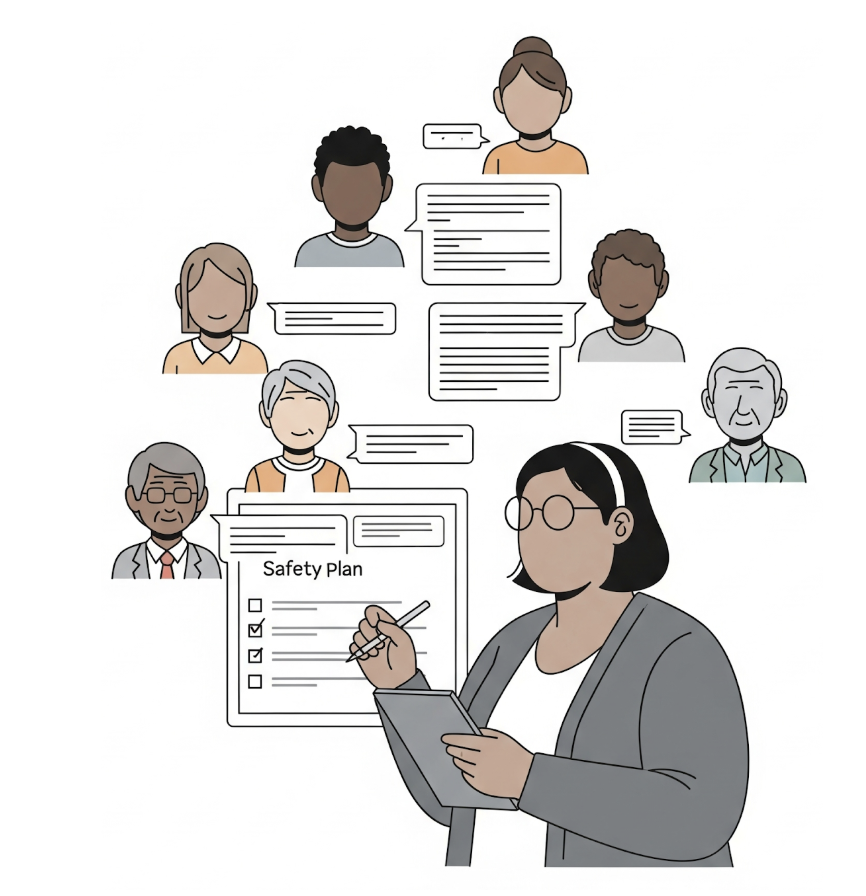}
  \caption{Generative AI enables the development of a diverse set of simulated clients experiencing a similar presenting concern (e.g., suicide risk), but having different cultural backgrounds or life experiences. These approaches give trainees opportunities to strengthen their cultural competence and practice intervention skills in more challenging or higher risk contexts without risk to human clients. This figure was made using generative AI.}
  \label{fig:myimage3}
\end{figure}

\section{Ability to provide in-the-moment feedback}

It is rare for clinicians to have access to in-the-moment support during clinical interactions. Relative to traditional supervision methods, live supervision and feedback (e.g., bug-in-the-ear and bug-in-the-eye technologies) are not feasible in therapy contexts \cite{Dufrene2018, Jones2017}. The rarity of these live supervision methods may be in part because it requires two people (trainee and clinician supervisor) for each patient interaction, they are and therefore are costly, and not easily scalable. Thus, they are rarely used, despite research finding that in-the-moment supervision and immediate feedback results in a greater trainee clinical competence and stronger therapeutic alliances between clinicians and their clients \cite{Weck2016, Vezer2021, Kluger1996}. 

Generative AI opens up possibilities for greater access to and potentially even more effective in-the-moment support for individuals providing mental health services. There is some emerging research evidence to support this use case. When practicing therapy skills with a patient simulation, for example, people who received in-the-moment AI generated feedback used significantly more reflective statements than those people who did not receive feedback \cite{Tanana2019}. Additionally, professional counselors perceive AI generated trainee support to be useful. In one study, generative AI was used to provide feedback to non-professional counselors’ drafted responses to depressed individuals. Professional counselors then rated these generative AI suggestions, finding them to accurately detect any inappropriate trainee responses (85-95 percent), suggest appropriate cognitive  strategies (79-84 percent), and provide constructive recommendations to the trainee (83-87 percent) \cite{Fu2023}. For more complex clinical problems, simulation and practice may lead to greater confidence and more rapid uptake of treatments after initial training \cite{Stade2024b}.

\begin{figure}[h!]
  \centering
  \includegraphics[width=0.8\textwidth]{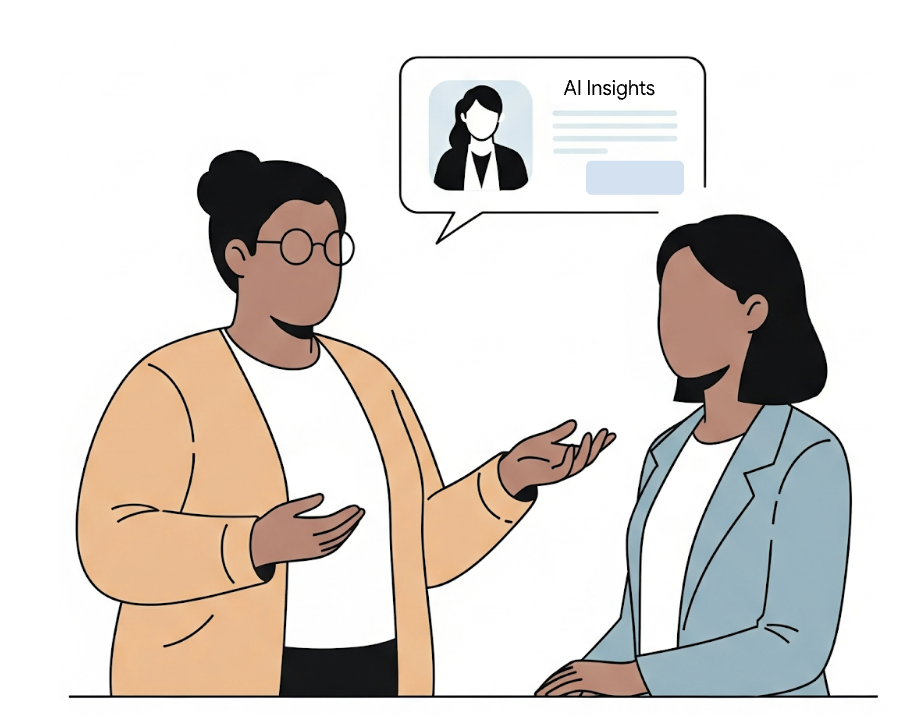}
  \caption{Generative AI allows for real time, in-the-moment feedback similar to bug-in-the-ear and bug-in-the-eye technologies. This figure was made using generative AI. }
  \label{fig:myimage4}
\end{figure}

\section{Ability to assess trainee strengths, areas for growth, and progress over time.}
\label{sec:headings}

Feedback and supervision are critical to acquiring skills \cite{Schwalbe2014}, and to improving and maintaining adherence with evidence-based practice \cite{Waller2009}. In order to provide trainees with this feedback, supervisors typically manually review video or audio recordings of trainee sessions with clients or rely on trainees’ retrospective self-report of how client interactions went. The former is time intensive, while the latter requires trainees’ accurate self-reflection and disclosure of challenges or areas in need of improvement to the supervisor. The vast majority of therapists believe they are well above average in therapeutic skills and they tend to report that their clients improve at rates far higher than would be expected \cite{Walfish2012, Parker2015}. These self-assessment biases make it unlikely that trainees consistently report on their own performance or how their client is progressing accurately. 

It is possible for generative AI to improve the quality of this supervision and feedback. Strengths of generative AI include its strong language understanding as well as its ability to synthesize large amounts of information. As such, it could be used to review transcripts or recordings of trainee sessions and summarize trainee strengths and weaknesses for the supervisor. Additional benefits could be to provide examples from a given session to illustrate those strengths and weaknesses, to evaluate changes in competence over time, and to point out specific segments of the recordings to the supervisor for their review. Using generative AI in these ways to support the supervisor has the potential to reduce time demands on supervisors, freeing up their time to supervise more trainees, as well as improve the quality of their supervision by giving supervisors the insights they need to provide accurate and objective feedback. Relatedly, generative AI could be used to measure clinicians ongoing alignment with evidence-based practice and to share objective feedback with the clinician or, if relevant, their supervisor or clinic leadership. More broadly, this information could be used to allocate refresher training and  incentivize provision of evidence-based care (e.g., from insurance) or could be used to help the public identify high quality clinicians.

\begin{figure}[h!]
  \centering
  \includegraphics[width=0.8\textwidth]{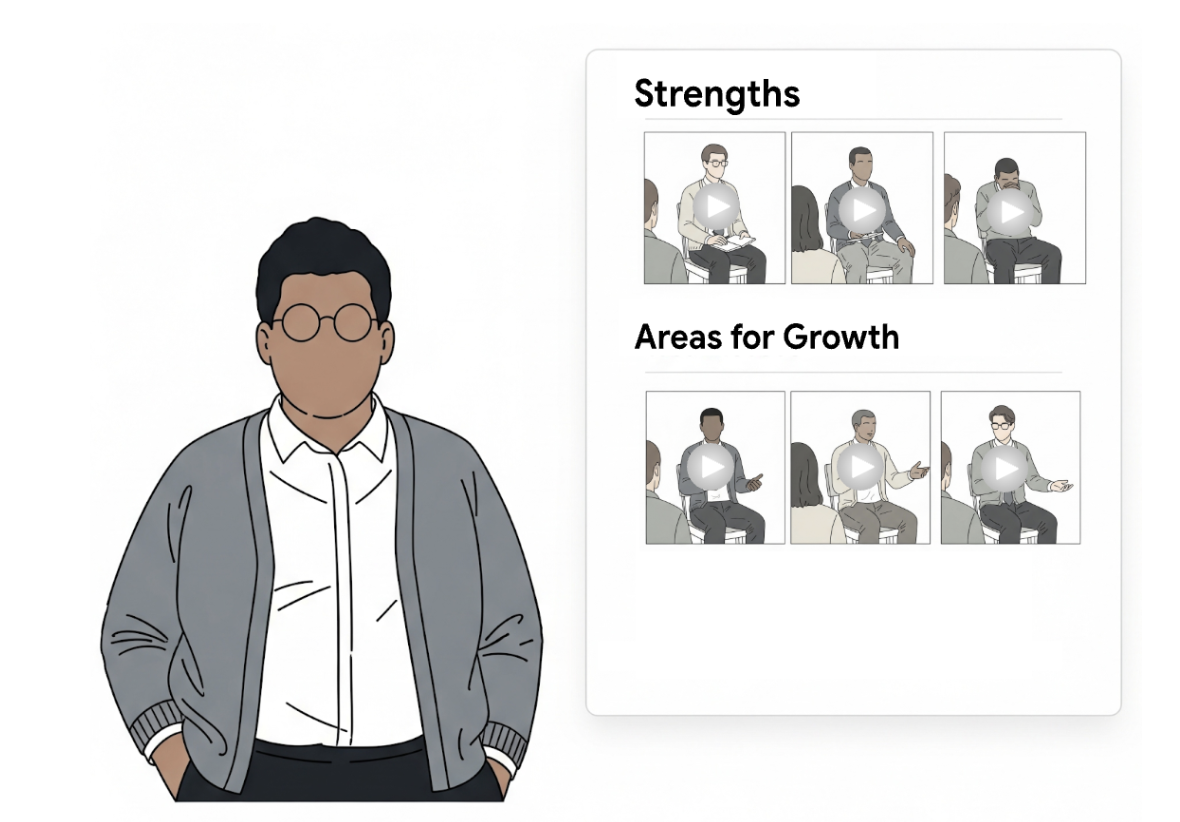}
  \caption{Generative AI could be used to summarize strengths and areas of improvement for trainees and their clinician supervisors, track changes in those areas over time, and identify specific examples (e.g., video clips) that represent those areas. This could not only save trainees and clinician supervisors the time needed to identify these strengths and weaknesses and scroll through video recordings of sessions to find examples, but ultimately may improve the quality of that supervision and clinical care. This figure was made using generative AI. }
  \label{fig:myimage5}
\end{figure}

\section{Lower risk}

Relative to therapist chatbots, using generative AI to support training in mental health service provision carries far fewer risks \cite{Lawrence2024}. When generative AI operates independently from humans in providing mental health support, risks are high and documented harm has been done. These harms include instances of models providing harmful and dangerous advice to people, increasing hopelessness, fostering dependence, and even encouraging suicidal behavior \cite{DeChoudhury2023}. Unlike for human healthcare clinicans, there are not currently ethical codes, comprehensive regulations, nor clear legal consequences should models not follow those codes and regulations.

Instead, human clinicians are obligated to adhere to professional ethics codes, standards of the profession, and legal obligations. As such, human-in-the-loop solutions that ensure that human experts retain control for decision making decreases the potential of harm. People still provide the direct support, but generative AI is there to augment their education and training.

\begin{figure}[h!]
  \centering
  \includegraphics[width=0.8\textwidth]{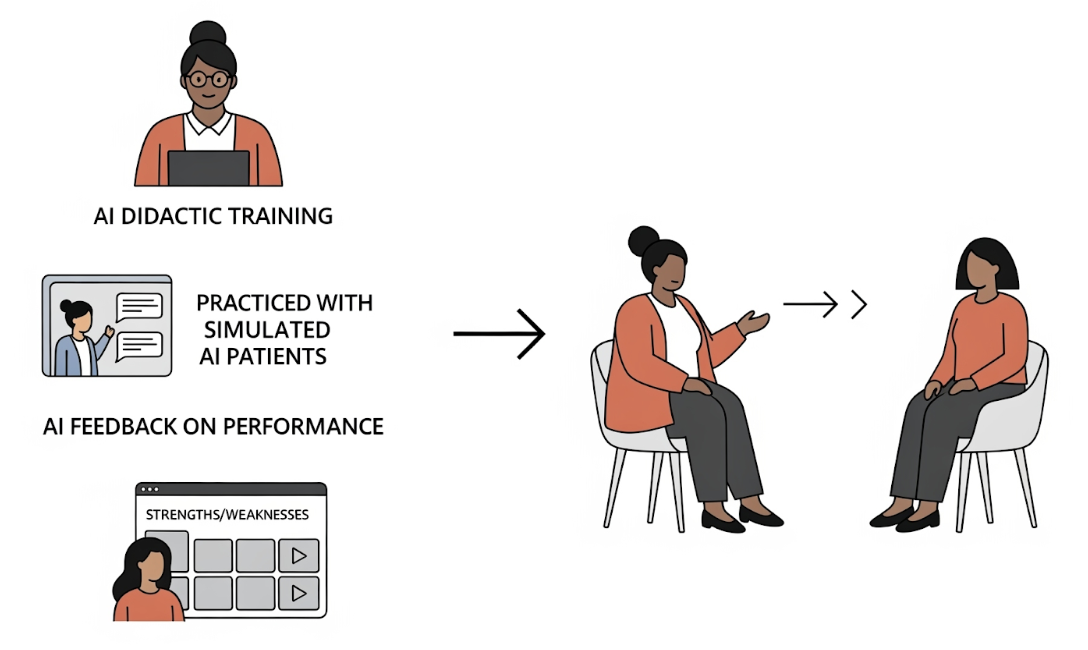}
  \caption{By focusing use of generative AI on trainee education, clients benefit from receiving mental health services from human clinicians, which carries far fewer risks than solutions where mental health services are delivered using AI without humans-in-the-loop. This figure was made using generative AI.}
  \label{fig:myimage6}
\end{figure}

\section{A Case Example: Using Generative AI to train veterans to have conversations about mental health}

Veterans die by suicide at rates 50

\begin{figure}[h!]
  \centering
  \includegraphics[width=0.8\textwidth]{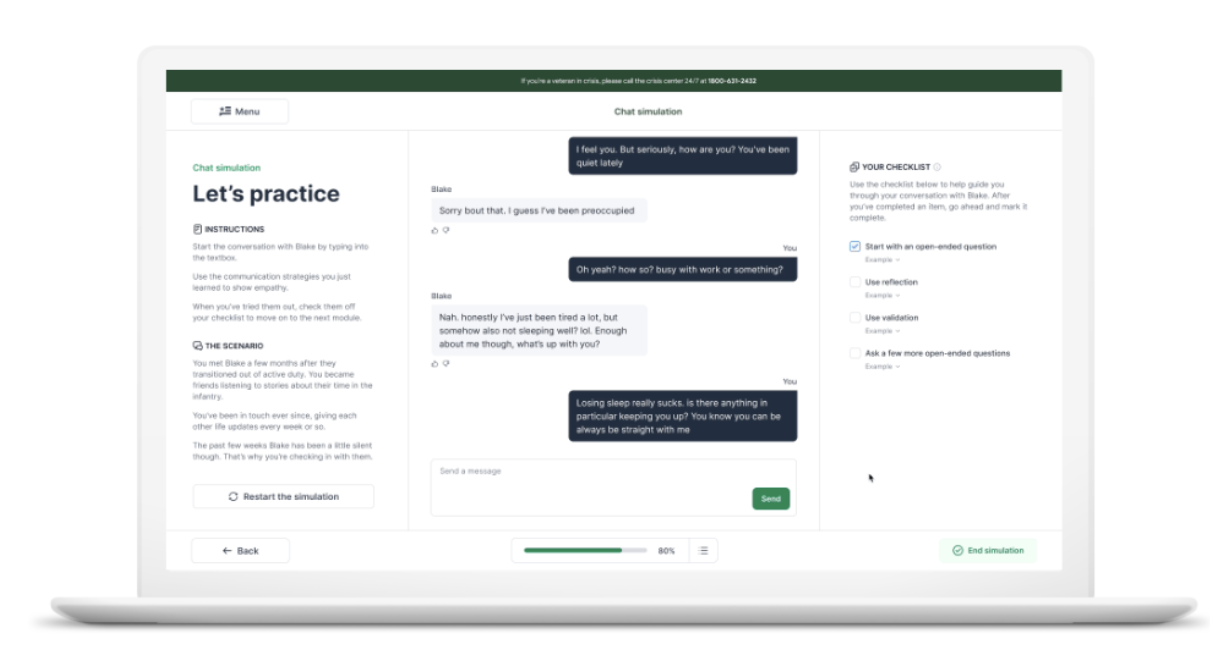}
  \caption{HomeTeam participants complete realistic simulations to practice skills learned in the content modules, such as building trust when talking about mental health and asking about suicide. They also engage in critical “Practice a Key Moment” simulations with immediate feedback on strengths and opportunities.}
  \label{fig:myimage7}
\end{figure}

Here, we highlight the ways in which HomeTeam illustrates the benefits of leveraging AI to support training in the provision of mental health support.

(1) Scalability: Since launch, thousands of veterans have practiced providing mental health support to other veterans using HomeTeam. Veterans across the US – and from every military branch – have engaged with the platform. There are few barriers to additional scaling within the US given the low costs associated with maintaining the platform now that it is established, though additional tailoring would be needed to adjust for other languages or changes in veteran needs over time.

(2) Opportunities for interactive practice: Following each 30-minute educational module, Veterans engage in simulated conversational practice with an AI-powered simulation, recognizing that interactive practice is key to learning and solidifying new skills. Evidencing the importance of this interactive practice, over 85

(3) Opportunities to practice with clients with diverse identities, backgrounds, and presenting concerns: Veterans have unique experiences and needs, and generally report feeling more comfortable sharing their mental health concerns with other veterans in a professional context \cite{Yeterian2023}. As such, it is critical for veterans to practice conversing with an AI agent that represents the veteran experience. Over 600 veterans from diverse backgrounds and with experiences across different branches of the military provided input into the development of HomeTeam. This allows other veterans to practice with a simulation that mimics the conversations that they may have with a real veteran peer.

(4) Ability to provide in-the-moment feedback: As part of HomeTeam’s simulated practice, Veterans receive feedback on key components of effective conversations about mental health. For example, each module within HomeTeam includes a keystone Practice a Key Moment exercise where participants understand the potential goals of that moment, engage in that critical moment of a simulated conversation, and receive instant feedback on their performance aligned with the clinically-informed goals.

(5) Ability to assess trainee strengths, areas for growth, and progress over time: Because participants in HomeTeam receive immediate feedback at various points within each module, the ability to learn is significantly enhanced compared with other learning modalities in which feedback is delayed or manual. Across more extended learning journeys in ReflexAI tools where larger data collection is possible, over 9 in 10 participants express satisfaction with the learning experience, highlighting key factors of increasing both their skills and confidence. 

(6) Lower risk: HomeTeam’s chat simulation offers a low risk environment for veterans to practice new conversational skills. This practice is especially crucial for more challenging conversations like those about suicide risk. Veterans can feel free to engage in these simulated conversations without the additional pressure and anxiety that comes with working with a real person in crisis. Additionally, no veterans with mental health concerns are impacted as other veterans engage in these conversations for the first time and start to build their competence and confidence.

\section{Conclusions}

The conversation around mental health and AI has largely focused on the role of therapist chatbots \cite{Vaidyam2019, Hua2024}. Direct provision of mental healthcare using generative AI brings with it risks, given the lack of human oversight and involvement. As such, therapist chatbots have understandably led to deep concern among mental health clinicians and the public alike \cite{Grabb2024}. A more conservative approach has been to leverage AI to support clinicians with documentation, allowing them to instead focus on direct patient care. Although this could certainly reduce burnout and potentially capacity in some regards, simulations suggest that increasing the efficiency and caseload of current mental health professionals is not likely to significantly reduce the workforce gap \cite{Bruckner2011}. Instead, using generative AI to enhance and scale training in mental health service provision is lower risk and has the potential to have a greater impact by increasing the number of people trained to deliver mental health interventions well. 

This approach does not come without challenges, however. Training tools need to be accurate, clinically appropriate, culturally responsive, and personalized to the trainee and client populations that the trainee will work with, all while avoiding stereotyping client groups. Moreover, concrete benchmarks for evaluating AI training tools need to be developed before they are used to train clinicians. Evaluating generative AI clients simulating a wide range of mental health concerns and geographical, biological, and cultural backgrounds is itself a challenging open question. Furthermore, prior to the proliferation of generative AI, there were criticisms of training with simulated clients, given that they would have to be simplified for the virtual environment \cite{Sharpless2009} and that interacting over a computer screen may not mimic real life clinical practice \cite{Sharpless2009}. Generative AI provides new opportunities for more realistic, complex patient simulations and more and more therapy is now provided virtually. As such, practicing in the virtual environment and with generative AI-powered simulated clients may now be a far closer approximation of real clinical practice. Nevertheless, there is a need for empirical evidence showing that use of generative AI training tools does in fact improve clinical competence. Developing tools with these capacities will require close collaboration and true partnership between clinical content experts and engineering teams throughout the development and evaluation phases. 

If these challenges are handled well, generative AI has the potential to support training in mental health service provision and truly expand access to high quality mental healthcare. By using generative AI, there is potential for this training to be cost effective, to reach underserved regions, and to scale. Given this potential, there is a need for investment in research, clinical, and industry endeavors that focus on this critical use case of generative AI in the mental health domain.

\section{Acknowledgments}

We acknowledge and thank John Hernandez, MD,; Maja Mataric, PhD; and Daniel McDuff, PhD, for reviewing and providing helpful feedback on this paper.

\section{Conflicts of Interest}

HRL, TY, and MJM are employees of Google and receive monetary compensation from Google and equity in Google's parent company, Alphabet. In addition, MJB is a shareholder in Meeno Technologies, Inc and The Orange Dot (Headspace Health).


\bibliographystyle{unsrtnat} 
\bibliography{references}  

\end{document}